\documentclass[siggraph,acmtog]{acmart}

\usepackage{subfigure}
\usepackage{caption}
\usepackage{booktabs}
\usepackage{multirow}

\AtBeginDocument{%
  \providecommand\BibTeX{{%
    \normalfont B\kern-0.5em{\scshape i\kern-0.25em b}\kern-0.8em\TeX}}}

\setcopyright{acmcopyright}
\copyrightyear{2022}
\acmYear{2022}
\acmDOI{https://doi.org/10.1145/3550454.3555468
}

\acmConference[Conference acronym 'XX]{Make sure to enter the correct
  conference title from your rights confirmation emai}{June 03--05,
  2018}{Woodstock, NY}
\acmBooktitle{Woodstock '18: ACM Symposium on Neural Gaze Detection,
 June 03--05, 2018, Woodstock, NY} 
\acmPrice{15.00}
\acmISBN{978-1-4503-XXXX-X/18/06}

\acmSubmissionID{papers\_290s1}

\setcopyright{acmlicensed}\acmJournal{TOG}
\acmYear{2022}\acmVolume{41}\acmNumber{6}\acmArticle{271}\acmMonth{12} \acmDOI{10.1145/3550454.3555468}

\begin{document}

\title{NeuralMarker: A Framework for Learning General Marker Correspondence}

\author{Zhaoyang Huang}
\authornote{Both authors contributed equally to this research.}
\affiliation{%
  \institution{The Chinese University of Hong Kong and NVIDIA}
  \country{China}
}
\email{drinkingcoder@link.cuhk.edu.hk}

\author{Xiaokun Pan}
\authornotemark[1]
\affiliation{%
  \institution{Zhejiang University}
    \country{China}
}
\email{panxkun@gmail.com}

\author{Weihong Pan}
\affiliation{%
  \institution{Zhejiang University}
  \country{China}
}
\email{22151254@zju.edu.cn}

\author{Weikang Bian}
\affiliation{%
  \institution{The Chinese University of Hong Kong}
    \country{China}
}
\email{wkbian@outlook.com}

\author{Yan Xu}
\affiliation{%
  \institution{The Chinese University of Hong Kong}
  \country{China}
}
\email{yanxu@link.cuhk.edu.hk}

\author{Ka Chun Cheung}
\affiliation{%
  \institution{NVIDIA}
  \country{China}
}
\email{chcheung@nvidia.com}

\author{Guofeng Zhang}
\authornote{Guofeng Zhang and Hongsheng Li are corresponding authors.}
\affiliation{%
  \institution{Zhejiang University}
  \country{China}
}
\email{zhangguofeng@zju.edu.cn}

\author{Hongsheng Li}
\authornotemark[2]
\affiliation{%
  \institution{The Chinese University of Hong Kong}
  \country{China}
}
\email{hsli@ee.cuhk.edu.hk}


\begin{abstract}
We tackle the problem of estimating correspondences from a general marker, such as a movie poster, to an image that captures such a marker.
Conventionally,
this problem is addressed by fitting a homography model based on sparse feature matching.
However, they are only able to handle plane-like markers and the sparse features do not sufficiently utilize appearance information.
In this paper, we propose a novel framework NeuralMarker, training a neural network estimating dense marker correspondences under various challenging conditions, such as marker deformation, harsh lighting, etc.
Deep learning has presented an excellent performance in correspondence learning once provided with sufficient training data.
However, annotating pixel-wise dense correspondence for training marker correspondence is too expensive.
We observe that the challenges of marker correspondence estimation come from two individual aspects: geometry variation and appearance variation.
We, therefore, design two components addressing these two challenges in NeuralMarker.
First, we create a synthetic dataset FlyingMarkers containing marker-image pairs with ground truth dense correspondences. 
By training with FlyingMarkers, the neural network is encouraged to capture various marker motions.
Second, 
we propose the novel Symmetric Epipolar Distance~(SED) loss, which enables learning dense correspondence from posed images. 
Learning with the SED loss and the cross-lighting posed images collected by Structure-from-Motion~(SfM), NeuralMarker is remarkably robust in harsh lighting environments and avoids synthetic image bias.
Besides, we also propose a novel marker correspondence evaluation method circumstancing annotations on real marker-image pairs and create a new benchmark.
We show that NeuralMarker significantly outperforms previous methods and enables new interesting applications, including Augmented Reality~(AR) and video editing.

\end{abstract}

\ccsdesc[500]{Computing methodologies~Vision for robotics}
\ccsdesc[500]{Computing methodologies~Visual inspection}
\ccsdesc[500]{Computing methodologies~Image processing}

\keywords{marker, correspondence, augmented reality, video editing}

\begin{teaserfigure}
\includegraphics[width=1.0\linewidth, trim={0mm 60mm 0mm 0mm}, clip]{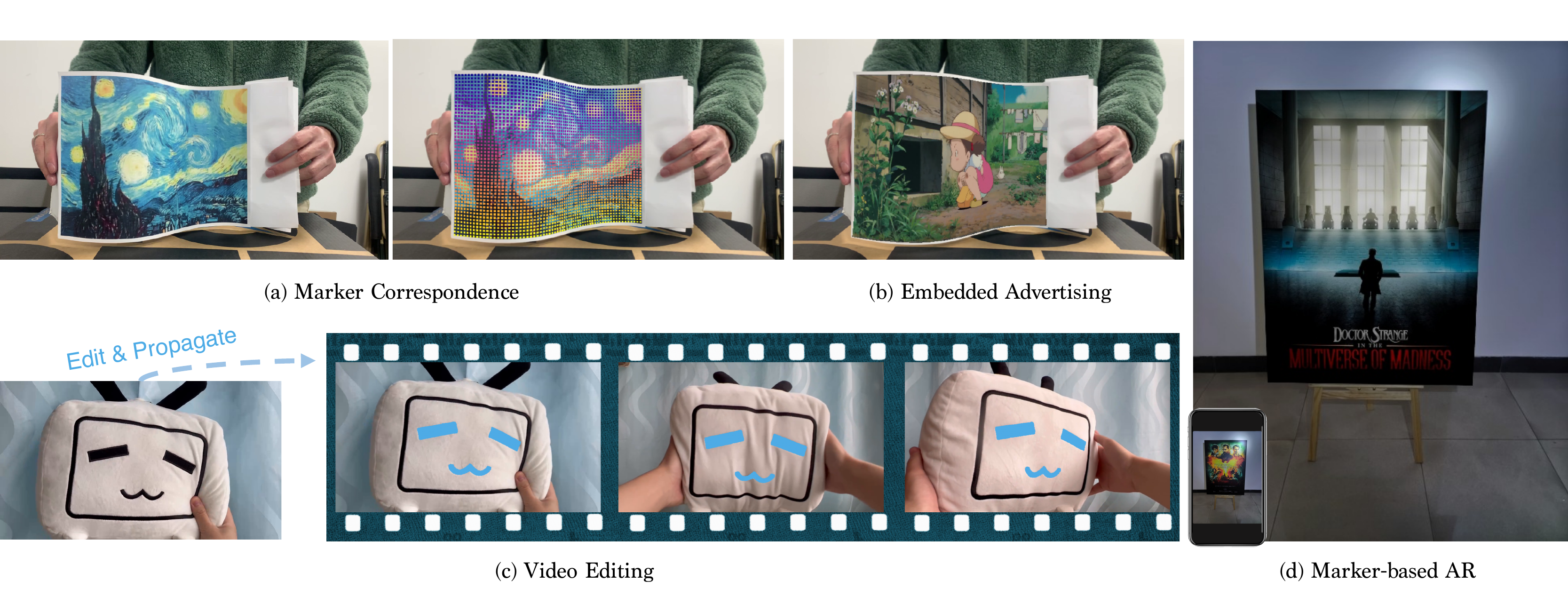}
  \caption{(a) The marker correspondence predicted by our NeuralMarker for an offhand marker. 
  (b) We can easily embed advertisement into movies and TV series via NeuralMarker.
  (c) We can edit a frame in a video clip and propagate the editing effects to the whole video clip.
  (d) The marker-based Augmented Reality~(AR). Please refer to the supplemented video for more results.}
  \label{fig:teaser}
\end{teaserfigure}

\maketitle
\section{Introduction}
General marker correspondence estimation aims at finding corresponding locations at a reference image for each pixel in the query marker.
In contrast to fiducial markers that are designed patterns, such as learned pattern~\cite{yaldiz2021deepformabletag,hu2019deep} and binary pattern~\cite{wang2016apriltag}, a general marker is an offhand arbitrary marker given by the user.
Without a strong prior of the pattern structure, general marker correspondence estimation is quite challenging.
Meanwhile, general marker correspondence serves as a core module in many downstream applications, such as marker-based Augmented Reality~(AR) and video editing.
Conventional general marker correspondence estimation is implemented by fitting a homography~\cite{szeliski2007image} with sparse features~\cite{lowe2004distinctive}.
However,
representing a marker or an image as a set of individual sparse features is ineffective,
and the homography model only supports an SE(3) transformation of a plane, which is unable to handle deformed markers.
In recent years, deep learning has presented an extraordinary performance in correspondence estimation~\cite{teed2020raft,huang2022flowformer} but training data~\cite{mayer2016large} is the foundation of the data-driven methods.
We propose a novel framework NeuralMarker, including a neural network, the training data preparation, and the corresponding loss functions, to learn to estimate pixel-wise marker correspondence.
The architecture design of our neural network follows the principle of RAFT~\cite{teed2020raft}, which computes the correlation of image features encoded by siamese image feature encoders as appearance similarity and infers pixel-wise correspondences from the appearance similarity via a motion regressor.
NeuralMarker directly estimates pixel-wise dense correspondences from the entire marker and image without the homography model, which fully utilizes the appearance information and gets rid of the plane constraints. 
However, annotating the ground truth dense correspondence for real marker-image pairs by humans is infeasible because the number of marker pixels is quite large.
We observe that the challenges in marker correspondence learning can be summarized as two aspects and are relative individual:
1) The geometry variations such as complex marker deformations. The motion regressor needs to capture various geometry variations in case of facing extrapolation in test scenarios.
2) The appearance variation such as lighting changes, requires the image feature encoder invariant to lighting so that we can obtain a robust appearance similarity pattern to support the correspondence estimation.
We thus design two components, each of them consisting of the training data with corresponding loss functions in our NeuralMarker to address these two challenges.

Motivated by FlyingChairs and FlyingThings~\cite{dosovitskiy2015flownet,mayer2016large} for optical flow estimation, we create a synthetic dataset FlyingMarkers for learning marker correspondence.
Given a marker and a reference background image,
we warp the marker according to a randomly generated geometric transformation and blend it with the reference background image as the synthesized reference image.
The marker, the geometric transformation, and the synthesized reference image constitute a training sample in FlyingMarkers.
Training with our FlyingMarkers makes the motion regressor encode sufficient geometric transformation priors.
However, in contrast to optical flow where consecutive images' appearance varies little, the reference image in practical scenarios may consist of significant appearance variations and they are difficult to synthesize.
Besides, the image feature encoder would be biased by synthesized images if it is only trained on synthesis images.
We thus propose the second component, learning from real images to cover various real appearance conditions.
We observe that Structure-from-Motion~\cite{schoenberger2016sfm} can collect real images covering various appearance conditions and compute their camera poses.
We propose the Symmetric Epipolar Distance~(SED) loss, which constrains the predicted correspondences based on camera poses.
Fueled by the SfM-collected images with ground-truth poses and supervised by the SED loss, the image feature encoder is able to learn robust features against appearance variations and avoids the synthesis-image bias.

Besides NeuralMarker, we also present a new benchmark to measure marker correspondence quality on real marker-image pairs in terms of marker deformation, viewpoint variation, and lighting variation, dubbed as DVL-Markers.
Similar to obtaining training data, building a marker correspondence benchmark is also faced with the difficulty of annotating ground-truth correspondences.
Actually, the estimated marker correspondence should be able to warp the original marker to align it to the captured marker in the reference image.
We, therefore, propose to measure the estimated marker correspondence via marker alignment consistency.

In this paper, we demonstrate 1) a synthetic dataset FlyingMarkers mimicking various marker motions for training and evaluating marker correspondence, 2) a SED loss with real posed images for training image feature encoders, which can be robust to appearance variation and removes the synthetic image bias, 3) a benchmark DVL-Markers to evaluate marker correspondence on real images, and 4) a series of interesting applications~(see Fig.~\ref{fig:teaser}).

\section{Related Work}
\textit{Marker Correspondence Estimation}
Marker correspondence estimation aims at estimating dense correspondences from a marker to a reference image that captures the marker.
This task can be divided into fiducial marker correspondence and general marker correspondence in terms of the type of used markers.
Fiducial marker methods~\cite{wang2016apriltag,olson2011apriltag,2015opencv,yaldiz2021deepformabletag,hu2019deep} are only interested in pre-defined markers, which are always binary patterns, and try to recover information encoded in the markers.
General marker methods~\cite{simonetti2013vuforia,sarosa2019developing} estimate dense correspondences for any markers given by users offhand instead of pre-defined markers.
Compared with fiducial markers, general markers do not require pre-arrangement of the scene so that they can be used in normal images and videos and support more interesting applications.
Traditional methods estimate correspondence for fiducial marker and general marker in a similar way, i.e., estimating marker pose via sparse correspondences~\cite{szeliski2007image}.
A series of works~\cite{narita2016dynamic,uchiyama2011deformable,degol2017chromatag,xu2011fourier,bencina2005improved} are devoted to designing better fiducial markers or utilizing marker prior more effectively.
Researchers also try to generate fiducial markers and estimate their correspondence through neural networks~\cite{grinchuk2016learnable,peace2021e2etag,yaldiz2021deepformabletag}, which significantly improves the accuracy and robustness.
However, to our best knowledge, no previous work addresses the general marker correspondence problem with deep neural networks.
Existing general marker correspondence estimation still follows the sparse feature matching, RANSAC~\cite{fischler1981random} with Homography, and Homography fitting pipeline~\cite{baker2006parameterizing,szeliski2007image}.
Although sparse features~\cite{detone2018superpoint,NEURIPS2019_3198dfd0,Dusmanu2019CVPR,sarlin2020superglue} have been improved by neural networks since SIFT~\cite{lowe2004distinctive}, this framework is still limited by the homography model, and sparse feature extraction is not a sufficient information encoding.
In this paper, we propose NeuralMarker, a novel framework for learning general marker correspondence estimation.
Compared with previous feature matching based methods,
NeuralMarker gets rid of the homography model and regresses correspondences from the whole marker and image, so it is able to handle deformed markers and obtains higher accuracy and robustness.

\textit{Learning to Estimate Dense Correspondence}
Optical flow is a mainstream dense correspondence estimation task, which concerns consecutive images in a video clip.
With abundant synthetic training data~\cite{dosovitskiy2015flownet,mayer2016large}, optical flow learning has achieved great success~\cite{teed2020raft,huang2022flowformer}.
FlowFormer~\cite{huang2022flowformer} pushes the accuracy further by utilizing transformers~\cite{vaswani2017attention,chu2021twins}.
Semantic correspondence~\cite{rocco2017convolutional,truong2020glu,shen2020ransac} is another genre that learns the dense correspondences of semantic objects between image pairs via data augmentation.
Truong et.al.~\cite{truong2021learning} achieves the state-of-the-art and extends the neural networks to learn geometric correspondences.
As most contents in images are static and rigid, epipolar constraints are widely employed in traditional optimization-based optical flow~\cite{valgaerts2008variational, wedel2009structure, yamaguchi2013robust}.
Some local feature learning methods leveraged epipolar constraints to train detectors~\cite{yang2019learning} and descriptors~\cite{wang2020learning}.
In NeuralMarker, we propose to use the Symmetric Epipolar Distance~(SED) as a weakly supervision loss.
Fueled by abundant cross-lighting posed images obtained from Structure-from-Motion~(SfM)~\cite{agarwal2011building,schoenberger2016sfm}, the image feature encoder learns robust features against lighting variations, which largely improves the following correspondence regression robustness.

\section{NeuralMarker}
Given a marker and a reference image that contains the marker, we propose a novel framework NeuralMarker, which trains a neural network identifying the marker's corresponding locations at the reference image for each pixel in the query marker. 
In this section, we will elaborate NeuralMarker in three aspects:
1) the neural network architecture,
2) supervised training with the synthetic dataset FlyingMarkers, and 3) weakly supervised training with real images and camera poses estimated by Structure-from-Motion~(SfM).

\subsection{Marker Correspondence Neural Network}

Intuitively, marker correspondence should be inferred from the appearance similarity between the query marker and the reference image, which is similar to the optical flow estimation principle.
We thus build our neural network architecture following RAFT, which can be viewed as two stages.
First, encoding feature maps from both marker and the reference image through a siamese image feature encoder and computing their correlations, which measures their appearance similarities.
To fully utilize the marker's information,
we use the pre-trained Twins-SVT~\cite{chu2021twins}, a transformer architecture encoding global features, as the image feature encoder, which has also been validated in the recent FlowFormer~\cite{huang2022flowformer}.
Second, iteratively regressing correspondence residuals with a motion regressor, which is a ConvGRU module~\cite{cho2014properties}, from the correlations and context features.

Optical flow only concerns temporally consecutive images, which generally share similar lighting and has small motion.
In contrast to optical flow, the marker usually undergoes large motion in the reference image, and the reference image may be captured in harsh lighting environments, which might cause its appearance to be significantly different from the marker.
We tackle these two challenges in the two stages of our neural network.
In the first stage, once the image feature encoder learned to encode lighting-invariant features, the correlations computed from such image features would remain constant regardless of the lighting variation of the reference image, and thus the motion regressor would not be affected.
We propose the SED loss to encourage our image feature encoder to be invariant to lighting variations.
In the second stage, the existence of large motion requires the motion regressor to be able to regress large displacement.
We thus propose FlyingMarkers, which provides synthesized reference images with various marker deformations to train a capable motion regressor.

\begin{figure}[t]
    \centering
    \includegraphics[width=1.0\linewidth, trim={10mm 100mm 100mm 0mm}]{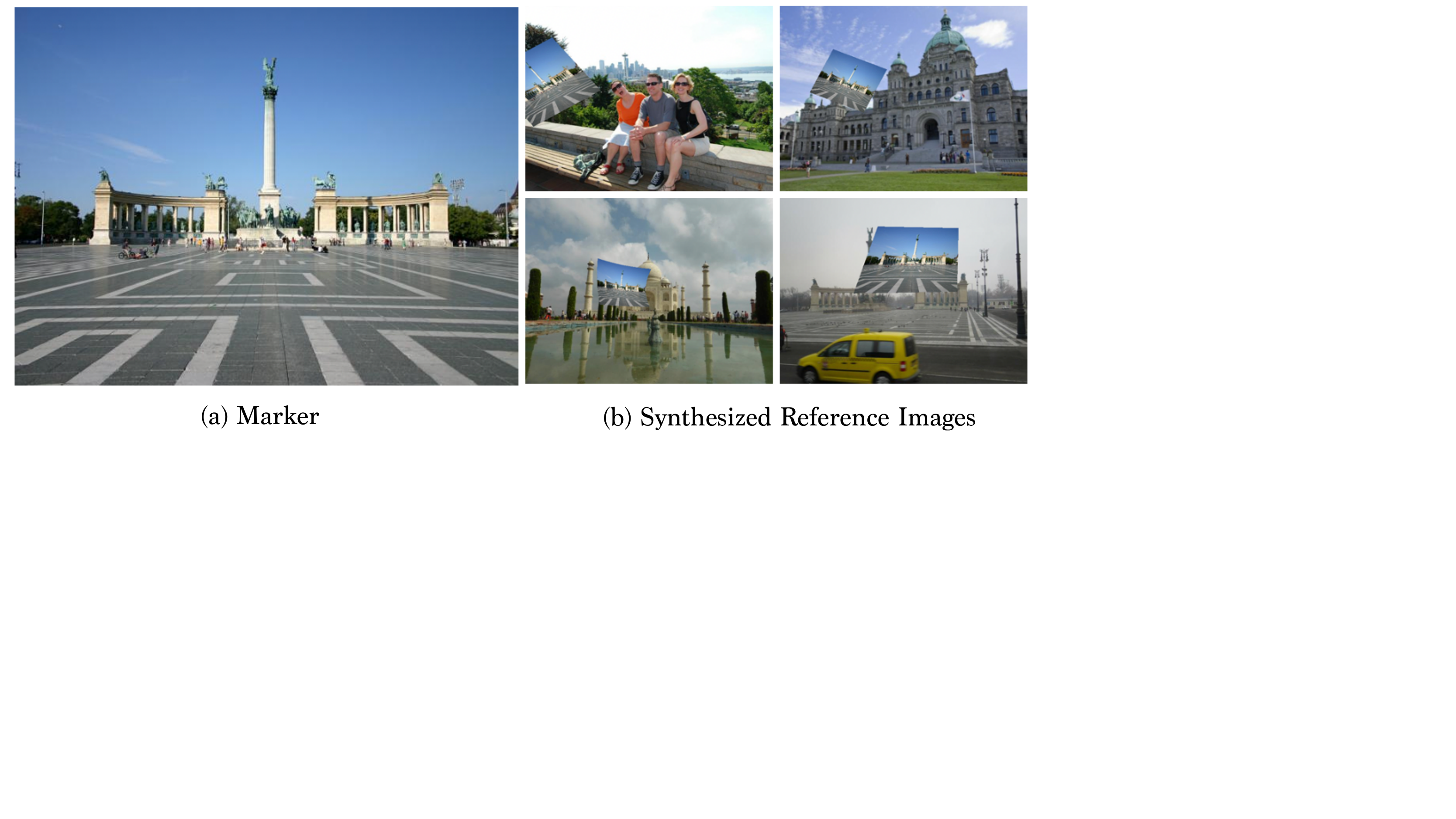}
        \caption{\label{Fig: flyingmarkers} FlyingMarkers. (a) A randomly selected marker. (b) Synthetic reference images of the marker (a).}
\end{figure}

\subsection{Supervised Training with FlyingMarkers\label{sec: flyingmarkers}}
It is challenging to collect the ground truth dense correspondences for marker-image pairs in real scenes and annotating such data by a human is infeasible.
Inspired by the success of FlyingChairs~\cite{dosovitskiy2015flownet} and FlyingThings~\cite{mayer2016large}, which are synthetic datasets for optical flow training, we propose FlyingMarkers, a synthetic dataset for marker correspondence training.
FlyingMarkers generates training data, including marker-image pairs with their ground truth dense correspondences, by synthesizing a marker deformed in an image.
Specifically, we select pairs of images from the MegaDepth dataset~\cite{li2018megadepth}. In each pair, one image is regarded as the marker and the other image as the background image.
We can synthesize a reference image that contains the marker via warping the marker and placing the marker in the background image.
Affine and homography transformations can fully represent geometric transformations of plane-like markers but cannot handle more complicated deformations.
Inspired by previous data augmentation techniques~\cite{melekhov2019dgc,truong2020glu}, we include a thin-plate spline~(TPS) model to synthesize the marker deformation.
We thus use the three following kinds of geometric transformations to warp markers:

\begin{itemize}
    \item \textit{Affine} transformation contains rotation, shear, and translation. We uniformly sample the rotation angle from -$\frac{\pi}{3}$ to $\frac{\pi}{3}$, the shear angle from -$\frac{\pi}{2}$ to $\frac{\pi}{2}$, and the translation from $0.75$ to $1.25$
    \item \textit{Homography} has 8 degrees of freedom~(DoF), which can be defined as the translation of four points from one image to another image. Therefore, we select the four corner points of the markers, randomly generate the four corners on the reference background image, and compute the homography matrix from the four-point translations as the randomly generated homography transformation.
    \item \textit{TPS}. We use a thin plate spline with 18 parameters, including 6 global affine motion parameters and 12 coefficients for controllable points. We also start from an identity mapping and randomly add a float number to each parameter from -0.5 to 0.5, where the pixel coordinates have been normalized to -1 to 1.
\end{itemize}

By computing the corresponding locations of the warped marker pixels, we obtain the dense correspondences from the marker to the reference image as ground truth.
We randomly sample a geometric transformation $\mathbf{T}$ from such three candidates, and directly supervise the neural network with the ground truth correspondences:

\begin{equation}
\begin{split}
L_{Syn}(I_M, I_R) = & \sum_{\mathbf{x}_i\in S}||f_{R\leftarrow M}(\mathbf{x}_i) - \mathbf{T}(\mathbf{x}_i)||_1. 
\end{split}
\end{equation}

$L_{Syn}$ is the loss used with the synthesized FlyingMarker dataset. Given the marker $I_M$ and the reference image $I_R$, our neural network predicts the correspondence $f_{R\leftarrow M}$ from $I_M$ to $I_R$ for all pixels $\mathbf{x}_i$ in the marker, where $S$ contains all pixels in $I_M$.
With the generated geometric transformation $\mathbf{T}$, we can compute their corresponding locations in the reference image $\mathbf{T}(\mathbf{x}_i)$, and supervise the predicted correspondences with L1 loss.
As the example shown in Fig.~\ref{Fig: flyingmarkers},
we can easily generate abundant marker-image pairs with proper supervision signals in this way.
FlyingMarkers contains 176,167 training samples, which cover various marker motions and deformations in total.
The motion regressor in our neural network is trained on FlyingMarkers sufficiently.

\subsection{Weakly Supervised Training with SfM Data}
\begin{figure}[t]
    \centering
    \includegraphics[width=1.0\linewidth, trim={10mm 60mm 200mm 0mm}]{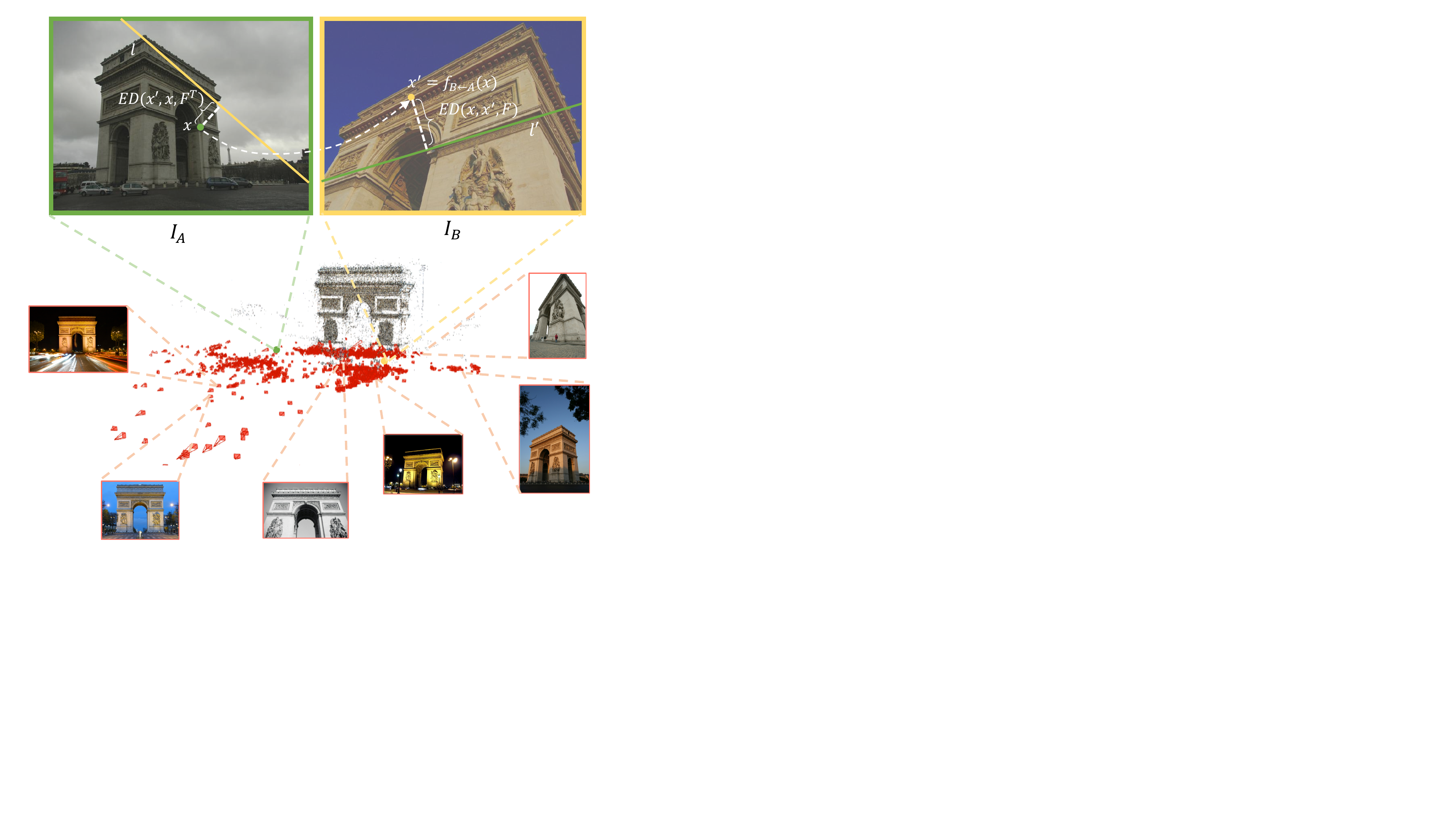}
        \caption{\label{Fig: sed-sfm} SED loss for SfM data. We compute camera poses with SfM for a collection of images that cover various lighting conditions. Given a pair of images with their relative camera poses, the SED loss constrains the location of a pixel location $x\in I_A$ to be on the corresponding epipolar line $l'$ (derived from the camera pose) in $I_B$, and vice versa.}
\end{figure}

Although FlyingMarkers provides abundant training data, it is synthetic and does not contain varying lighting marker-image pairs. 
We propose a symmetric epipolar distance (SED) loss to address this limitation.
As shown in Fig.~\ref{Fig: sed-sfm}, 
we can collect real images that cover various lighting conditions and compute their camera poses with SfM.
The proposed SED loss can train our neural network with such real and natural lighting-varying images.
Specifically,
we compute the fundamental matrix $\mathbf{F}$ from an image $I_A$ to another image $I_B$ according to their camera intrinsic parameters and relative camera pose from SfM.
$\mathbf{F}$ restricts that a pixel location $\mathbf{x}\in \mathbb{R}^2$ in image $I_A$ can only be mapped to one of the points on a line $l'=\mathbf{Fx}$, referred to as the {\it epipolar line} in image $I_B$.
For each point $\mathbf{x}$ in image $I_A$, our network estimates its corresponding point in image $I_B$ as $\mathbf{x}'=f_{B\leftarrow A}(\mathbf{x})$.
If the $I_A$-to-$I_B$ correspondences are ideal, the distance of the corresponding pixel location $x'$ to the epipolar line $l'$, named epipolar distance~(ED), shall be zero.
Reversely, $x$ is also supposed to lie on the epipolar line $l$ derived from $x'$ in image $I_B$ if the reverse flows are ideal, and the distance from $x$ to the epipolar line $l$ should also be zero.
The sum of the two epipolar distances is defined as the SED.
According to the epipolar geometry, the inverted fundamental matrix equals the transpose of the fundamental matrix, we can therefore compute the SED loss as
\begin{equation}
SED(\mathbf{x}, \mathbf{x}', \mathbf{F})
= ED(\mathbf{x}, \mathbf{x}', \mathbf{F}) + ED(\mathbf{x}', \mathbf{x}, \mathbf{F}^T).
\end{equation}

Given the $I_A$-to-$I_B$ dense correspondences $f_{B\leftarrow A}$ estimated by our neural network, we define the following SED loss to evaluate their accuracies by computing SED for all correspondences in $f_{B\leftarrow A}$,
\begin{equation}
L_{SED}(I_A, I_B)
= \sum_{\mathbf{x}_i \in S}SED(\mathbf{x}_i, f_{B\leftarrow A}(\mathbf{x}_i),\mathbf{F}),
\end{equation}
where $S$ is the set containing all pixel locations in $I_A$.
The proposed SED loss is only derived from the fundamental matrix (or relative camera pose). Therefore, the SED loss works even when there exist significant lighting variations between the pair of images.
The SED loss coupled with the SfM data serving as weak supervision effectively mitigates bias of the synthesis image and constant lighting.

\subsection{Training Neural Network}

As the markers used in FlyingMarkers come from MegaDepth dataset~\cite{li2018megadepth},
we can identify other images in MegaDepth surrounding the images that are used as markers.
For each training sample, including a marker $I_M$ and a synthesized reference image $I_{R1}$, in FlyingMarkers, we sample another image $I_{R2}$ that has common visible observations with $I_M$ in the MegaDepth dataset.
With such a triplet, we can train our neural network with both the supervised loss and the SED loss,

\begin{equation}
L_{all}(I_M, I_{R1}, I_{R2}) = L_{Syn}(I_M, I_{R1}) + L_{SED}(I_M, I_{R2}).
\end{equation}

We use a learning rate of $10^{-4}$, a weight decay of $5\times 10^{-5}$, the one-cycle learning rate scheduler, 12 recurrent iterations, a batch size of 12, 640$\times$480 image size, and 100k training iterations.

\begin{figure}[t]
    \centering
    \includegraphics[width=1.0\linewidth, trim={5mm 60mm 110mm 0mm}, clip]{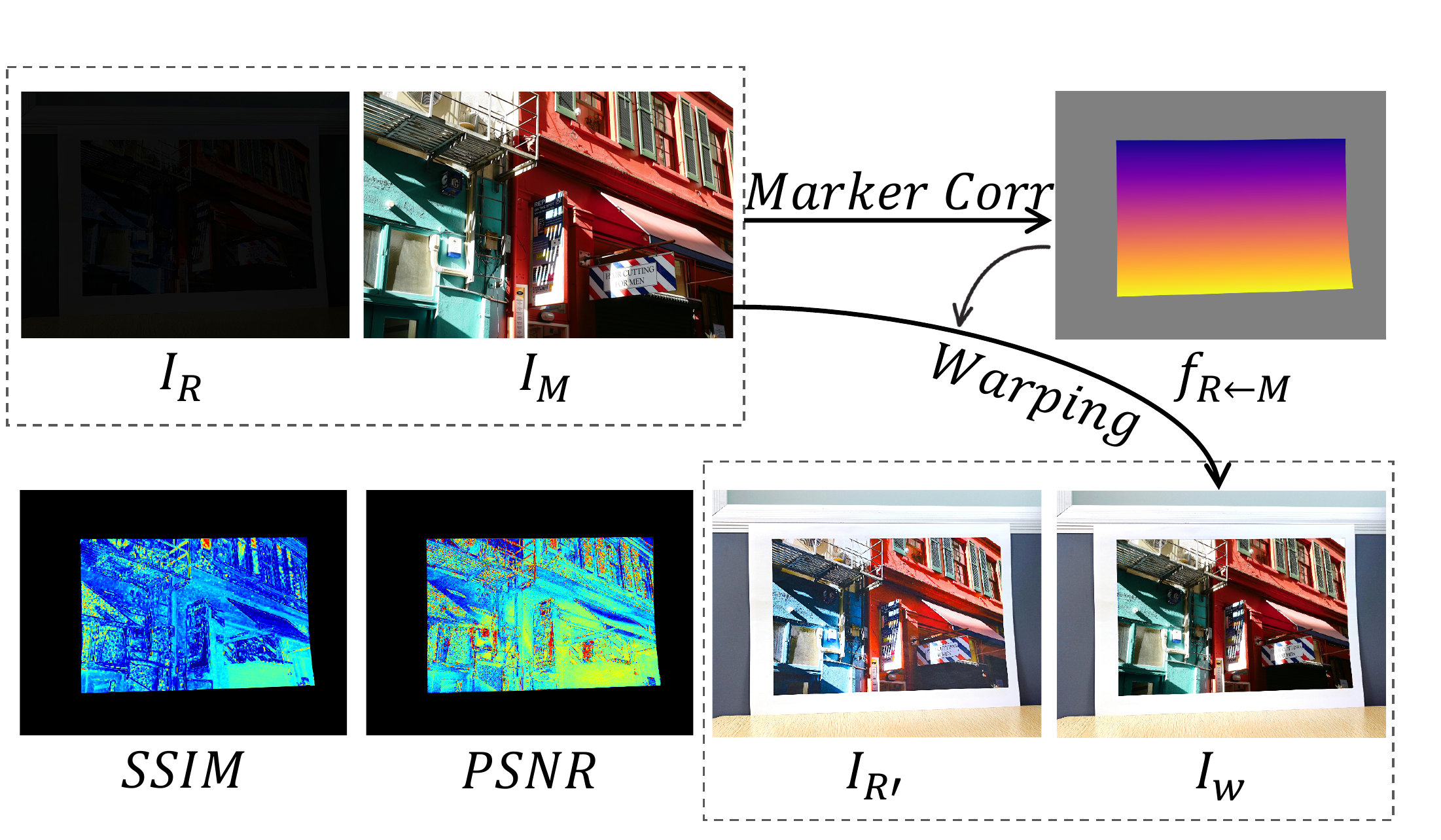}
        \caption{\label{Fig: dvl evaluation example}DVL-Markers Benchmark.
        Given the predicted correspondence $f_{R\leftarrow M}$ from the marker $I_M$ to the reference image $I_R$, evaluate it through the SSIM and PSNR metrics between the ground-truth image $I_{R'}$ and the synthesis image $I_w$ that is warped from $I_M$.
        }
\end{figure}

\section{Marker Correspondence Benchmark}

In this section, we introduce two benchmarks, including FlyingMarkers and DVL-Markers, and corresponding metrics to evaluate marker correspondence quality.

\subsection{FlyingMarkers}
Following the image synthesis and ground-truth marker correspondence generation introduced in Sec.~\ref{sec: flyingmarkers},
we create a test set to evaluate marker correspondence.
Given the estimated marker correspondence $f_{R\leftarrow M}$ and the ground-truth transformation $\mathbf T$, we can compute the End-Point Error~(EPE) for each marker pixel as
$EPE(\mathbf{x}_i) = ||f_{R\leftarrow M}(\mathbf{x}_i) - \mathbf{T}(\mathbf{x}_i)||_2$. 
In line with other dense correspondence evaluation~\cite{truong2021learning}, we employ the Percentage of Correct Keypoints~(PCK) metrics.
PCK-$\delta$ is the percentage of marker pixels $\hat{\mathbf{x}}_i$ whose correspondence EPE is smaller than a given threshold $\delta$.

\subsection{DVL-Markers Benchmark}

\begin{figure*}[t]
    \centering
    \includegraphics[width=1.0\linewidth, trim={10mm 50mm 0mm 0mm}, clip]{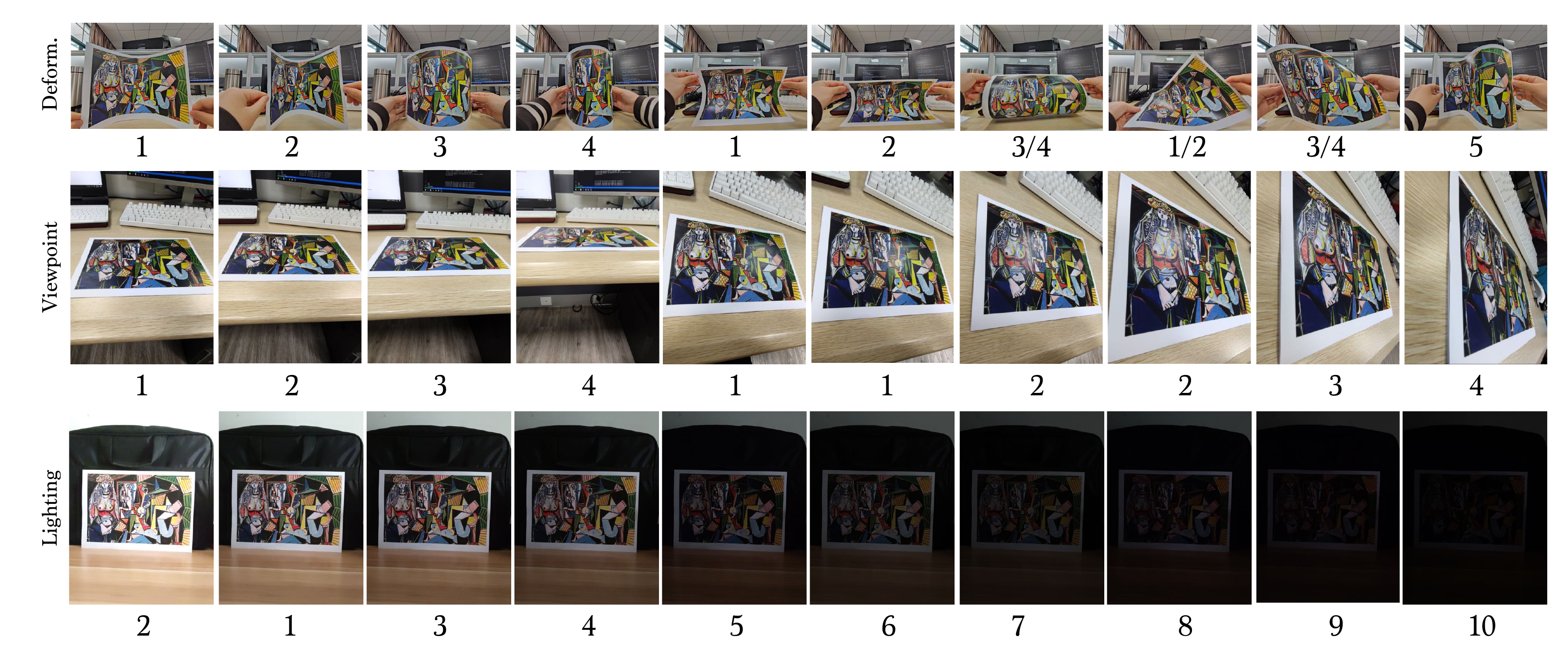}
        \caption{\label{Fig: test images} Test images in DVL-Markers. Each image is assigned a difficulty level. The 3/4 label means the image will be labeled as 3 or 4 according to the deformation degree.
        }
\end{figure*}

Existing benchmarks for optical flow~\cite{geiger2013vision,butler2012naturalistic} only evaluate correspondences between consecutive images in a video clip, which have small motions and similar lighting conditions.
Although FlyingMarkers provides quantitative marker correspondence evaluation, it still has two significant limitations: 1) the reference image is synthesized, and 2) the warped marker has the same lighting conditions as the original marker.
What if we would like to evaluate marker correspondence estimation with real images and varying lighting conditions?
The challenge of creating such a benchmark is similar to the problem in the training data generation: it is difficult to annotate pixel-wise correspondences for marker-image pairs, especially when there are marker deformations and challenging lighting.
We observe that if the estimated marker correspondences are of high quality, the marker warped according to the marker correspondences should be well aligned to the marker captured in the image.
We, therefore, tackle the challenge by evaluating the correspondences according to marker alignment consistency.
We propose a new benchmark, DVL-Markers, containing marker-image pairs for marker correspondence evaluation and the images are pictures taken in real scenes.
DVL-Markers contain three sets: deformation, viewpoint, and lighting, respectively, standing for challenging cases of marker deformation, viewpoint variation, and harsh lighting. 
Specifically,
we warp the marker $I_M$ with the estimated correspondences $f_{R\leftarrow M}$, denoted as $I_w$~(Fig.~\ref{Fig: dvl evaluation example}).

\begin{equation}
\begin{split}
I_w & = warp(I_M, f_{R\leftarrow M}), \\
S' & = \{\mathbf{x}_i| \mathbf{x}_i \in S, I_w(\mathbf{x}_i) \neq (0,0,0) \}. 
\end{split}
\end{equation}

The warped marker is assumed to be consistent with the image content inside the covered area $S'$.
Structural Similarity~(SSIM) and Peak Signal-to-Noise Ratio~(PSNR) are classical metrics evaluating image reconstruction quality.
We thus compute the marker alignment quality of such marker-image pair, $SSIM_I$ and $PSNR_I$, which are the average of SSIM and PSNR for all valid pixels $\mathbf{x}_i\in S'$ between the warped marker $I_w$ and the reference image $I_R$:

\begin{equation}
\begin{split}
SSIM_I & = \frac{1}{||S'||}\sum_{\mathbf{x}_i\in S'}SSIM(I_w(\mathbf{x}_i),I_R(\mathbf{x}_i)), \\
 PSNR_I & =  \frac{1}{||S'||}\sum_{\mathbf{x}_i\in S'}PSNR(I_w(\mathbf{x}_i), I_R(\mathbf{x}_i)).
\end{split}
\end{equation}

This strategy is effective for evaluating marker-image pairs of viewpoint variation and deformation but is unable to work on pairs of images with lighting variation because of the nature of SSIM and PSNR metrics.
Therefore, for each picture $I_R$ took under low-lighting in the lighting set, we take another picture $I_{R'}$ using the same camera pose and with abundant lighting.
We still feed the low-light image $I_R$ to correspondence estimation methods but compute the SSIM and PSNR metric with the well-lit image $I_{R'}$.

\subsection{Difficulty Levels in DVL-Markers}

In DVL-Markers, we prepare 10 markers and take 10 pictures for each marker under each condition, which consists of 300 test images.
We show the test images of one test marker in Fig.~\ref{Fig: test images}.
Each row contains test images of a subset and we further divide the deformation, viewpoint, and lighting subsets into 5, 4, and 10 difficulty levels. Each test image is assigned a difficulty level.

In the deformation subset, we take 2 horizontally concave, 2 horizontally convex, 2 vertically concave, 1 vertically convex, 1 diagonally concave, 1 diagonally convex, and 1 wave-like deformation, which corresponds to test images in columns 1-10 of row 1.
We observe that convex deformation is more difficult than concave deformation, so small and large concave deformations are labeled as 1 and 2, and small and large convex deformations are labeled as 3 and 4. The wave-like deformation is the most challenging case, which is labeled as 5. Test images in columns 7, 8, and 9 are labeled as two different levels according to the deformation degree.

In the viewpoint subset, we take test images according to the angle between the camera viewpoint and the marker’s negative normal. We take test images at around 10°, 25°, 50°, and 75°, which are labeled from 1 to 4.

In the lighting subset, we control the lighting condition by turning on/off the light and adjusting the camera shutter. We set 10 illumination levels ranging from underexposure to overexposure.
The image in column 2 is regarded as a well-lit image, which is labeled as level 1. The overexposed image in column 1 is labeled as level 2. Then the images from column 3 to column 10 are captured under decreasing lighting, so the difficulty level gradually increases. 

\begin{figure*}
    \centering
    \includegraphics[width=1.0\linewidth, trim={10mm 60mm 50mm 0mm}, clip]{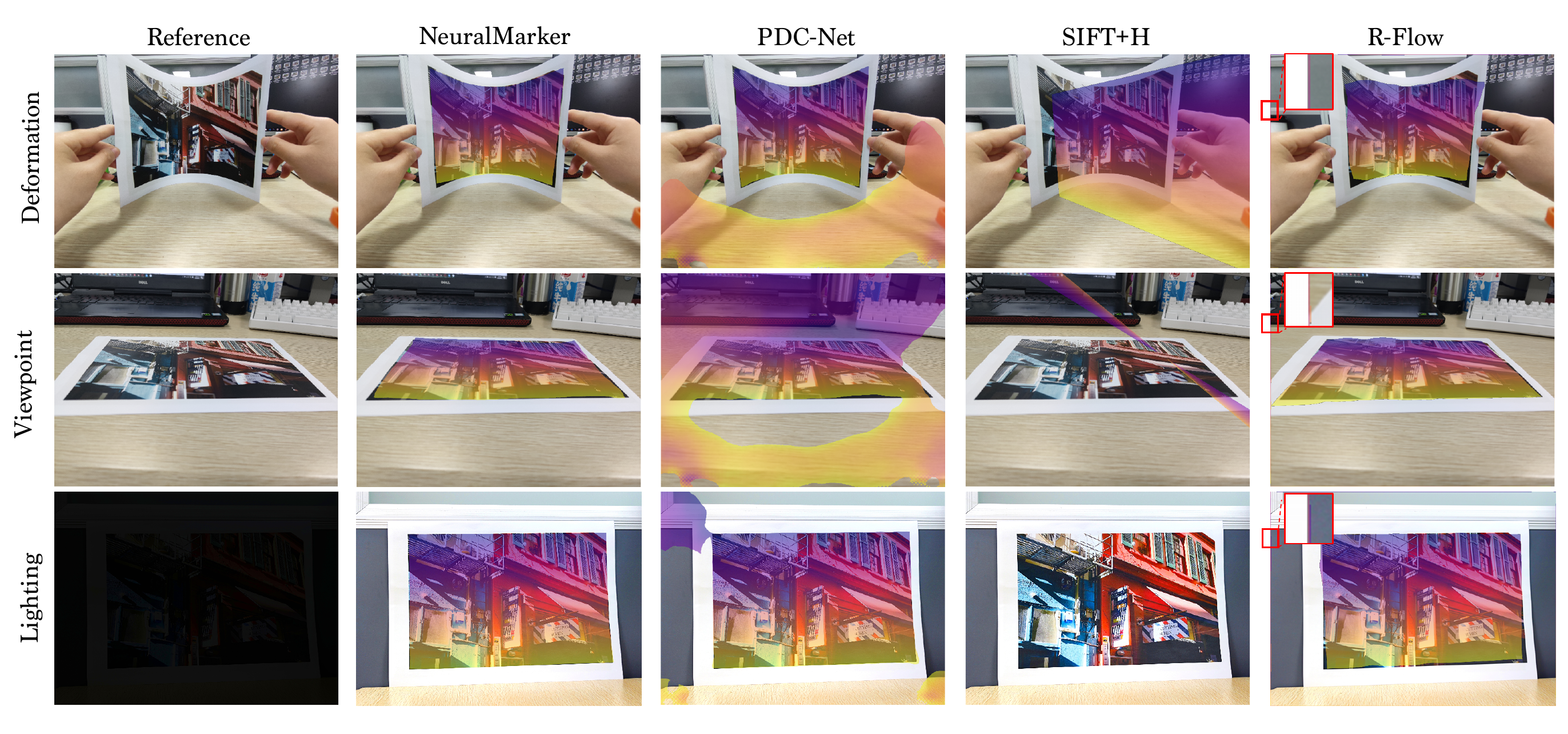}
    \caption{Marker Correspondence Visualization on the DVL-Markers Benchmark. We show an extreme reference image of the same marker for each condition and visualize the estimated marker correspondences. 
    }
    \label{fig: dvl-markers comparison}
\end{figure*}

\section{Experiments}

\begin{table*}[t]
    \centering
    \begin{tabular}{cccccccccc}
    \toprule
      \multirow[c]{2}{*}{Method} &  \multicolumn{3}{c}{Deformation}  & \multicolumn{3}{c}{Viewpoint}& \multicolumn{3}{c}{Lighting}\\
      \cmidrule(r{1.0ex}){2-4}\cmidrule(r{1.0ex}){5-7}\cmidrule(r{1.0ex}){8-10}
     & SSIM$\uparrow$ & PSNR$\uparrow$ & Failed$\downarrow$ & SSIM$\uparrow$ & PSNR$\uparrow$ & Failed$\downarrow$ & SSIM$\uparrow$ & PSNR$\uparrow$ & Failed$\downarrow$ \\
     \hline 
     SIFT+H~\cite{lowe2004distinctive} & 0.27/0.25 & 10.97/10.68  & 0\% & 0.46/0.57 & 12.59/13.21 & 0\% & 0.64/0.58  & 13.94/12.80 & 25\%\\
     SP+SG+H~\cite{sarlin2020superglue} & 0.33/0.29 & 11.36/10.93 & 12\% & 0.66/0.63 & 14.53/13.89  & 19\% & 0.70/0.67 & 14.34/13.52 & 21\% \\
     R-Flow~\cite{shen2020ransac} & 0.32/0.30 & 11.42/10.89 & 0\% & 0.42/0.52 & 12.04/12.18 & 0\% & 0.68/0.77  &  14.63/14.96 & 0\% \\
     PDC~\cite{truong2021learning} & 0.24/0.23 & 10.46/10.27 & 0\% & 0.38/0.32 & 11.40/11.07 & 0\% & 0.46/0.46 & 11.59/11.47 & 0\%\\
     NeuralMarker~(Ours)  & 0.65/0.69 & 14.38/14.28 & 0\% & 0.70/0.77 & 15.20/15.96 & 0\%  & 0.79/0.82 & 16.29/15.66 & 0\%\\
     \bottomrule
    \end{tabular}
    \caption{Evaluation on the DVL-Markers benchmark. We evaluate marker correspondence methods by warping markers according to predicted correspondences and compute the image consistency. We use SSIM (mean/median) and PSNR (mean/median) as the image consistency metric. NeuralMarker obtains extraordinary accuracy and robustness compared with all other methods.
    }
    \label{tab: dvl-marker}
\end{table*}

We conduct a series of experiments and provide results to evaluate the marker correspondence accuracy. 
To our best knowledge, we are the first that focuses on the general marker correspondence problem with deep neural networks.
Previous marker correspondences are derived from sparse feature matching-based homography fitting.
We thus select SIFT~\cite{lowe2004distinctive} and SuperPoint~\cite{detone2018superpoint} as sparse features for homgography fitting as the competitive counterparts.
Based on SuperPoint, SuperGLUE~\cite{sarlin2020superglue} is a state-of-the-art feature point matcher so we adopt it and denote this combination as SP+SG.
‘+H' denotes that the matched sparse features are used to fit a homography model.
We also select two general correspondence estimation methods: RANSAC-Flow~\cite{shen2020ransac} and PDC-Net~\cite{truong2021learning} for comparison.
PCK metric requires ground-truth pixel-wise correspondences, which are unable to be collected in real scenes. To evaluate the correspondence quality in real scenes, we warp the marker according to the estimated marker correspondence and compute the marker alignment quality, i.e., SSIM and PSNR.
Therefore, we use PCK on FlyingMarkers, a synthetic dataset, and SSIM and PSNR on DVL-Markers, a real scene dataset.

\begin{table}[t]
    \centering
    \begin{tabular}{ccccc}
    \bottomrule
    Method & PCK-1 & PCK-3 & PCK-5\\
     \hline 
     SIFT+H~\cite{lowe2004distinctive} & 0.57 & 0.71 & 0.74\\
     SP+SG+H~\cite{sarlin2020superglue} & 0.38 & 0.63 & 0.70 &\\
     R-Flow~\cite{shen2020ransac} & 0.10 & 0.42  & 0.48 \\
     PDC-Net~\cite{truong2021learning} & 0.75 & 0.82 & 0.82\\
     NeuralMarker~(Ours)  & 0.89 & 0.99 & 0.99  \\
    \bottomrule
    \end{tabular}
    \caption{Evaluation on the test set of FlyingMarkers with PCK-1, PCK-3, and PCK-5.
    }
    \label{tab: marker}
\end{table}

\textit{FlyingMarkers.} Our NeuralMarker obtains superior performance and outperforms compared methods on the FlyingMarkers test set.
Note that the PCK-5 of NeuralMarker achieves 99\%, which denotes that the motion regressor in our neural network is able to capture almost all markers' transformations and deformations.
PDC-Net presents consistently inferior performance and its precision from PCK-1 to PCK-5 does not significantly vary, which denotes that PDC-Net totally fails in these cases rather than lacks accuracy.

\textit{DVL-Markers Benchmark.}
As shown in Tab.~\ref{tab: dvl-marker}, We compute the mean and median of $SSIM_I$ and $PSNR_I$.
Note that homography fitting requires at least 4 correspondences, so homography-based methods may fail when valid feature matches are less than 4. 
We thus also present the failure percentage and compute their mean SSIM and PSNR from valid data.
NeuralMarker shows dominant superiority compared with all previous methods across all of the three sets.
In the Deformation set, all methods present inferior performance except NeuralMarker.
Although RANSAC-Flow and PDC-Net do not use a homography model, they do not show significant advantages over SIFT+H and SP+SG+H.
In the Viewpoint set,
SIFT+H and SP+SG+H show better performance compared with other methods because this set conforms to the homography model that they used, but our NeuralMarker still significantly outperforms them even without the homography model.
In the Lighting set,
SIFT+H does not obtain reasonable performance.
The 25\% failure ratio denotes its vulnerability to lighting variation.
The other learning-based methods are consistently better than SIFT+H because the learned features are more robust than SIFT features against lighting variations.
The most competitive method is RANSAC-Flow, but it is still inferior compared to our NeuralMarker and does not obtain good performance under the other two conditions. 
We also qualitatively compare marker correspondences in Fig.~\ref{fig: dvl-markers comparison}.
SIFT+H is limited by the homography model and fails in the low-lighting image.
PDC-Net predicts many chaotic correspondences, which will severely impact downstream applications.
RANSAC-Flow consistently presents correspondence leakage over the boundary of the image.
Please zoom in to see the details.
Marker correspondences estimated by NeuralMarker are coherent and accurate.

\begin{figure*}
    \centering
    \includegraphics[width=1.0\linewidth, trim={0mm 75mm 0mm 0mm}, clip]{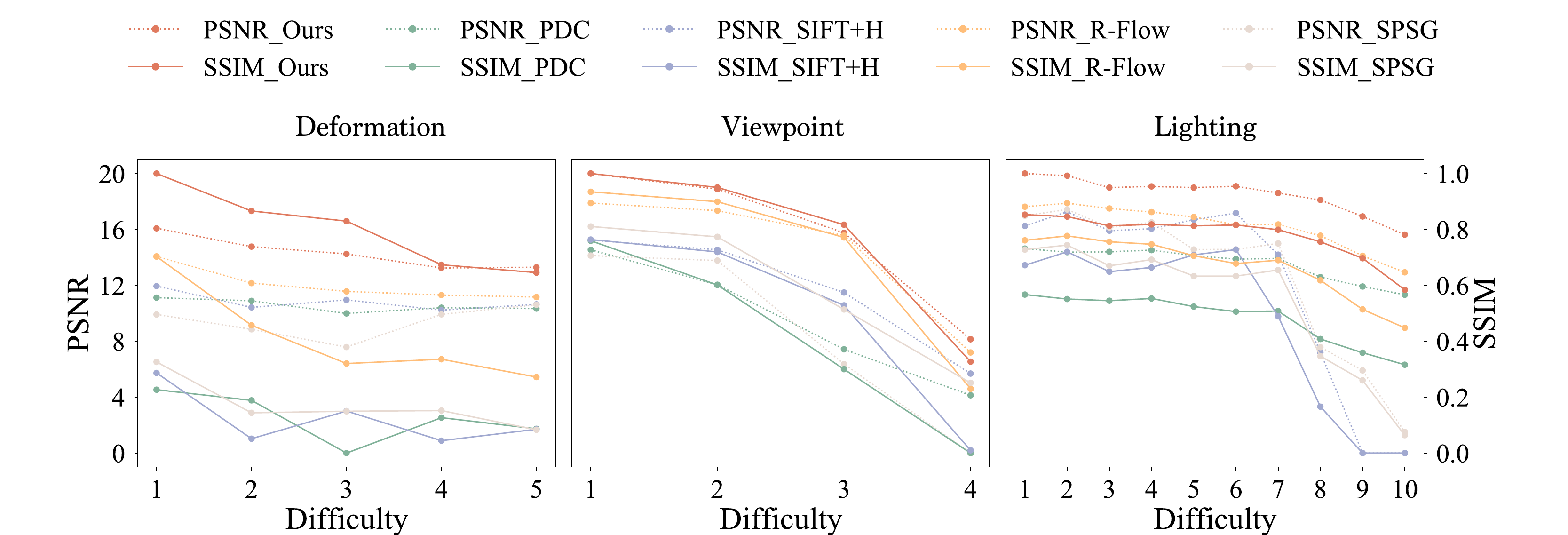}
    \caption{Performance in increasing difficulty levels on DVL-Markers.
    }
    \label{fig: dvl-markers performance analysis}
\end{figure*}

\textit{Performance in increasing difficulty levels}~(Fig.~\ref{fig: dvl-markers performance analysis}). 
We further divide the deformation, viewpoint, and lighting subsets into 5, 4, and 10 difficulty levels according to deformation degree, viewpoint angle, and lighting intensity.
Our NeuralMarker presents extraordinary performance, especially in the Deformation and Lighting subset.
In the deformation set, the PDC-Net and the SIFT+H rarely achieve 12 PSNR or 0.4 SSIM. which denotes that they are fragile when the marker is deformed.
SIFT+H fails in the lighting subset when the difficulty level equals 9 and 10 because there are no sufficient features for homography fitting.
The performance of the other two competitors is closest to NeuralMarker in the Viewpoint set but still presents an evident gap.

\begin{table}[t]
    \centering
    \begin{tabular}{cccccc}
    \toprule
     Scene num.  & 1 & 10 & 50 & 100 & 176\\
     \hline 
    PCK-1 & 0.042 & 0.335 & 0.799  & 0.848 & 0.887 \\
    PCK-3 & 0.259 & 0.818 & 0.976 & 0.985 & 0.989\\
    PCK-5 & 0.453 & 0.915 & 0.991 & 0.994 & 0.996\\
    \bottomrule
    \end{tabular}
    \caption{PCK with different scales of training data. The models are evaluated on the test set of FlyingMarkers.
    }
    \label{tab: training data scale}
\end{table}

\textit{PCK with different scales of training data}. In FlyingMarkers, we obtain 176 scenes from the MegaDepth and extract 1k images in each scene for training.
To present the performance of models that are trained by different scales of data, we now train our neural network with images from 1, 10, 50, 100, and 176 scenes and evaluate the models on the test set of FlyingMarkers. As shown in Tab.~\ref{tab: training data scale}, our NeuralMarker achieves the best performance when using images from all scenes.

\textit{Ablation Study.} 
We use the PCK-1 on FlyingMarkers and the median of SSIM on DVL-Markers in the ablation study.
We start from the baseline `CNN + $L_{Syn}$', which uses the CNN image feature encoder following RAFT, then replaces the CNN with the pre-trained Twins-SVT (`Twins + $L_{Syn}$'), and finally adds the SED loss (`Twins + $L_{Syn}$ + $L_{SED}$').
The baseline shows good performance on the synthetic data but can not be generalized to real-world data.
Twins-SVT improves the generalization but is still unsatisfactory.
The SED loss sacrifices little accuracy on the synthetic data for much better generalization in the real world.

\begin{table}[t]
    \centering
    \begin{tabular}{ccccc}
    \toprule
      & FlyingMarkers & \multicolumn{3}{c}{DVL-Markers}\\
     \cmidrule(r{1.0ex}){2-2}\cmidrule(r{1.0ex}){3-5}& PCK-1 & D & V & L \\
     \hline 
    CNN + $L_{Syn}$ & 0.95 & -0.02 & 0.01 & 0.06 \\
    Twins + $L_{Syn}$ & 0.95 & 0.52 & 0.39 & 0.46 \\
    Twins + $L_{Syn}$ + $L_{SED}$ & 0.89 & 0.69 & 0.77 & 0.82 \\
    \bottomrule
    \end{tabular}
    \caption{Ablation study. `Twins + $L_{Syn}$ + $L_{SED}$' is the final model we use.
    }
    \label{tab: dvl-marker}
\end{table}

\section{Applications}

\begin{figure}[t]
    \centering
    \includegraphics[width=1.0\linewidth, trim={5mm 115mm 150mm 0mm}, clip]{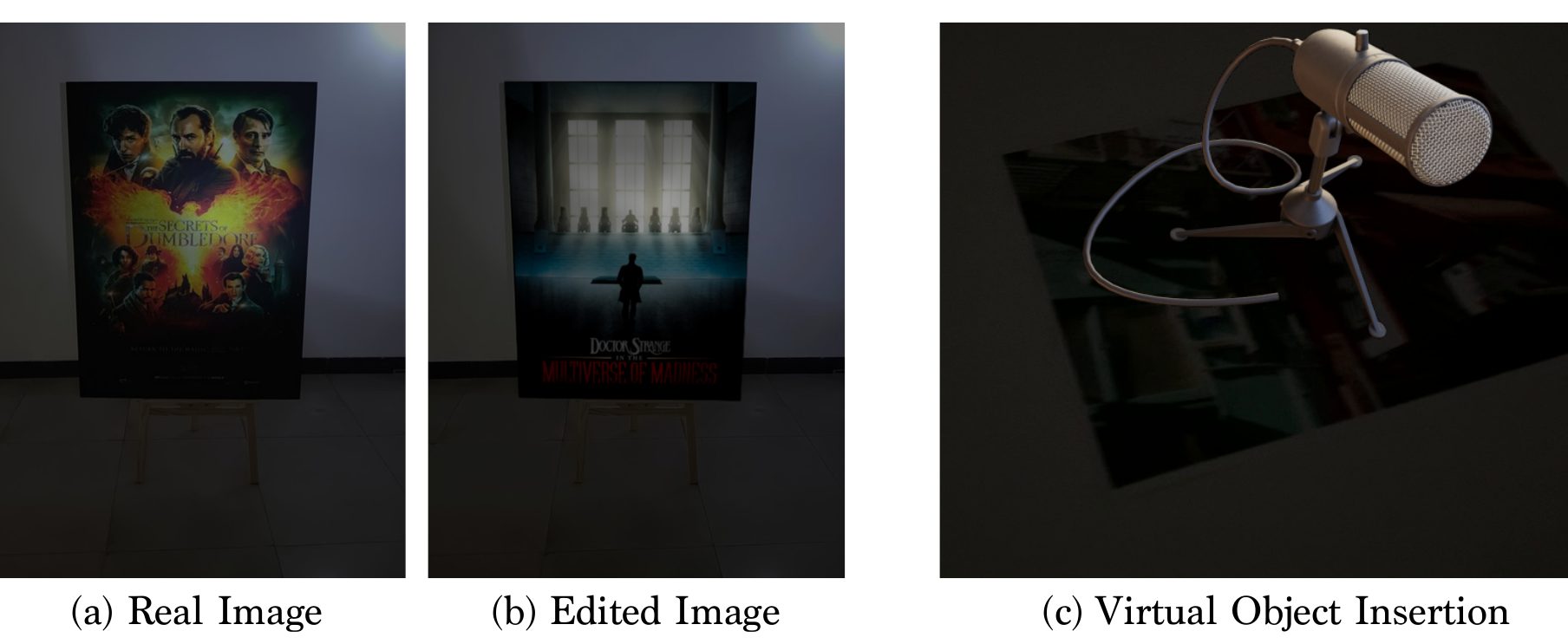}
        \caption{\label{Fig: harsh lighting} AR in harsh lighting environments.
        }
\end{figure}

Compared with fiducial markers, a general marker does not need to pre-arrange the environment, which makes NeuralMarker capable of processing general videos, such as live streaming, movies, and TV series.
For example, advertising in TV series with the guidance of marker correspondences in Fig.~\ref{fig:teaser} (b).
Besides, even in homemade videos, we can use an elegant marker that can be recognized by humans rather than unmeaning binary patterns.

Compared with previous sparse feature-based general marker correspondence estimation, NeuralMarker demonstrates superiority in robustness and accuracy and supports deformed markers.
Marker-based AR is one of the mainstream AR applications in many commercial AR systems~\cite{simonetti2013vuforia}.
These functionalities that are extended by our NeuralMarker can enable more interesting AR effects.
In Fig.~\ref{Fig: harsh lighting}, we present that we can realize AR effects in harsh lighting environments according to marker correspondence predicted by NeuralMarker.
We replace the reflectance of the poster in the real image (a) guided by marker correspondence and preserve the shading through NIID-Net~\cite{luo2020niid}.
We also identify the marker in such a low-light environment and insert a virtual object on the marker plane.
Besides, NeuralMarker also supports editing contents on a deformed marker.
A common requirement in video editing is to add some effects to an object and expect the editing results to propagate over the video clip while maintaining consistency in the content.
GANGealing~\cite{peebles2021gan} reveals that we can edit a template image and propagate the results to the related images through predicted correspondences.
However, the template image and the correspondence learning in GANGealing require a pre-trained generative model for each target object.
In contrast, our NeuralMarker is able to predict correspondences for an offhand given marker.
We therefore can extract one frame from the video clip as the marker, edit the marker, and then propagate the editing effects to the whole video clip guided by marker correspondences. Please refer to the supplemented video for more results.

\section{Limitations}
\begin{figure}[t]
    \centering
    \includegraphics[width=1.0\linewidth, trim={10mm 115mm 110mm 0mm}, clip]{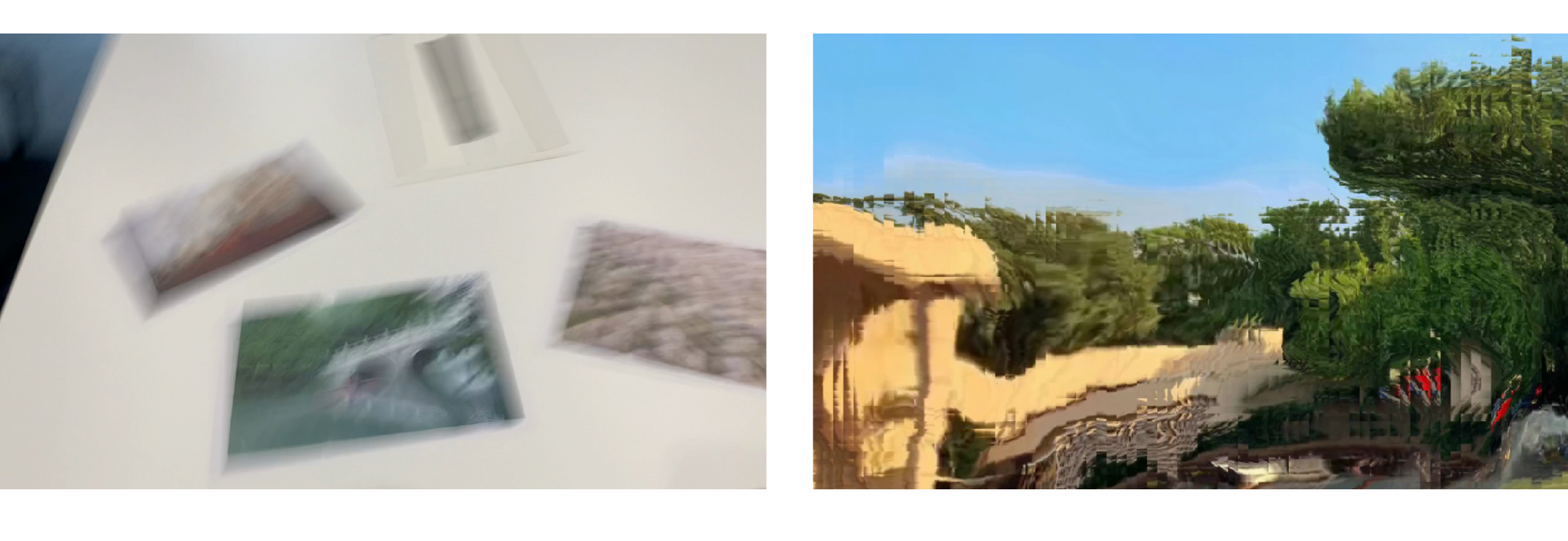}
    \includegraphics[width=1.0\linewidth, trim={0mm 95mm 110mm 0mm}, clip]{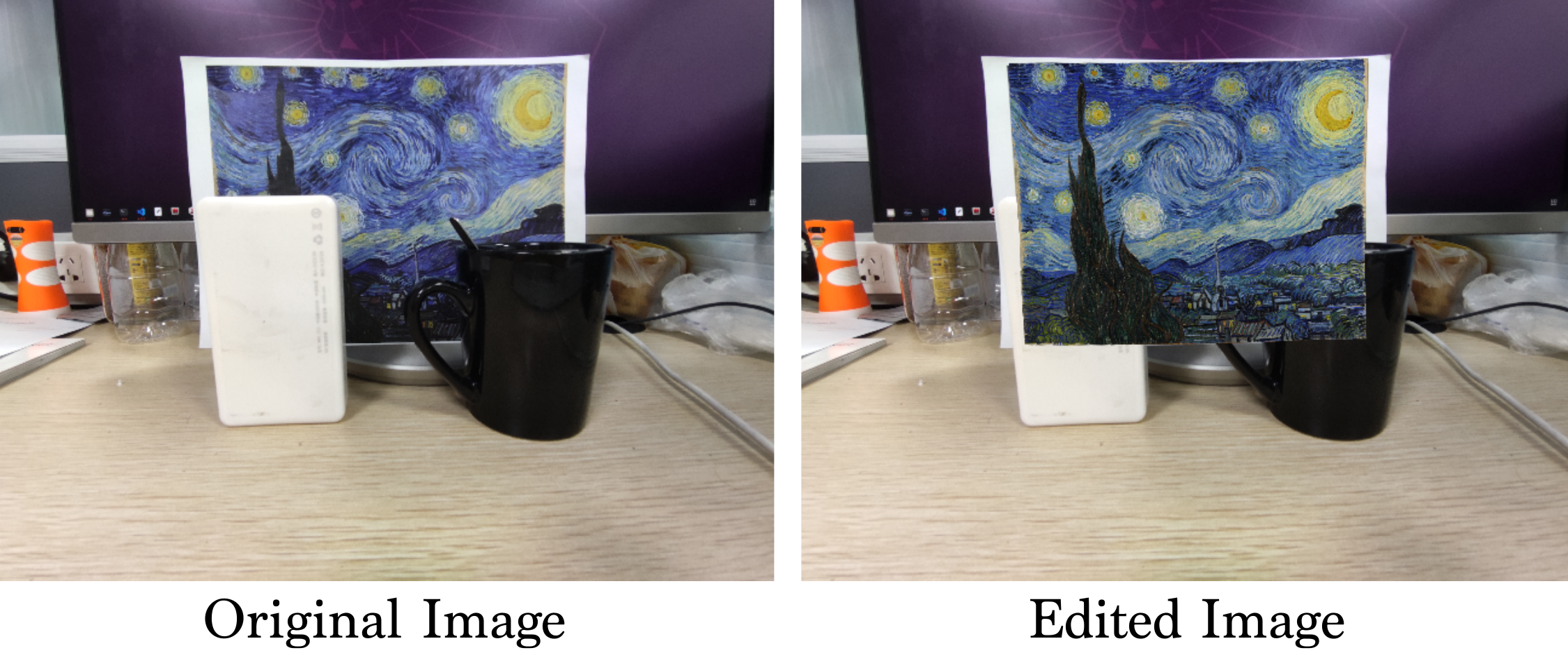}
        \caption{\label{Fig: limitations} Limitations. Row 1: NeuralMarker still fails when the motion blur is so severe. Row 2: Without occlusion mask prediction, the marker directly warped by predicted correspondences will cover the occluder.
        }
\end{figure}

NeuralMarker cannot simultaneously estimate marker correspondence for multiple markers.
Besides, NeuralMarker does not estimate the occluded regions, so directly editing the images according to estimated marker correspondences will cover the occluding object.
In the future, we can design a one-shot marker detector to crop each marker for the following accurate marker correspondence estimation, which enables multi-marker correspondence estimation.
To avoid affecting the occluder in front of the marker when editing the image according to marker correspondences, we can augment the FlyingMarker dataset by randomly inserting occluders and additionally learn to predict the occlusion mask. The predicted mask can assist in image editing.
The DVL-Markers benchmark does not quantitatively evaluate the performance for test images with motion blur. We qualitatively compare NeuralMarker with other methods in the supplemented video. Although NeuralMarker presents the best robustness but still fails when the motion blur is so severe.
We show two failed image editing cases in Fig.~\ref{Fig: limitations}.

\section{Conclusion}
We have developed NeuralMarker, a novel framework for learning general marker correspondence.
To tackle the challenges of lacking training image pairs with realistic lighting and deformation variations, we propose the novel Symmetric Epipolar Distance loss to train image pairs with only ground-truth relative camera poses.
NeuralMarker significantly outperforms previous general marker correspondence methods on both synthetic and real-world data.
Based on the marker correspondence predicted by NeuralMarker, some interesting but challenging applications can be realized now, such as AR in harsh lighting environments and video editing.

\section{Acknowledgements}
We thank Rensen Xu, Yijin Li and Jundan Luo for their help.
Hongsheng Li is also a Principal Investigator of Centre for
Perceptual and Interactive Intelligence Limited (CPII).
This work is supported in part by CPII, in part by the General Research Fund through the Research Grants Council of Hong Kong under Grants (Nos. 14204021, 14207319) and in part by ZJU-SenseTime Joint Lab of 3D Vision.

\bibliographystyle{ACM-Reference-Format}
\bibliography{egbib.bib}

\appendix
\end{document}